\documentclass{llncs}
\pdfoutput=1
\usepackage{makeidx} 
\usepackage{listings}
\usepackage{graphicx}
\usepackage{booktabs}
\usepackage{CJKutf8}

\let\llncssubparagraph\subparagraph
\let\subparagraph\paragraph
\usepackage[compact]{titlesec}
\let\subparagraph\llncssubparagraph
 
\begin{document}
\mainmatter             

\title{Cross-Lingual Predicate Mapping Between Linked Data Ontologies}
\titlerunning{Cross-lingual Predicate Mapping Between Linked Data Ontologies} 

\author{Gautam Singh\inst{1}, Saemi Jang\inst{2}, Mun Y. Yi\inst{2}}
\authorrunning{Gautam Singh et al.} 
\tocauthor{Gautam Singh, Sammy Jang, Mun Y. Yi}

\institute{
Department of Electronics and Electrical Engineering,\\Indian Institute of Technology Guwahati, Kamrup, Assam, India\\
\email{gautam.singh@iitg.ernet.in}
\and
Department of Knowledge Service Engineering,\\Korea Advanced Institute of Science and Technology, \\
291 Daehak-ro, Yuseong-gu, Daejeon, Republic of Korea\\
\email{sammy1221@kaist.ac.kr}, \email{munyi@kaist.ac.kr}
}

\maketitle              

\begin{abstract}
Ontologies in different natural languages often differ in quality in terms of richness of schema or richness of internal links. This difference is markedly visible when comparing a rich English language ontology with a non-English language counterpart. Discovering alignment between them is a useful endeavor as it serves as a starting point in bridging the disparity. In particular, our work is motivated by the absence of inter-language links for predicates in the localised versions of DBpedia. In this paper, we propose and demonstrate an ad-hoc system to find possible \emph{owl:equivalentProperty} links between predicates in ontologies of different natural languages. We seek to achieve this mapping by using pre-existing inter-language links of the resources connected by the given predicate. Thus, our methodology stresses on semantic similarity rather than lexical. Moreover, through an evaluation, we show that our system is capable of outperforming a baseline system that is similar to the one used in recent OAEI campaigns.
\end{abstract}

\section{Introduction}
Ontologies in different natural languages often differ in quality in terms of richness of schema or richness of internal links. Bizer et al. \cite{bizer} refers to various ontology-related problems like maintaining quality, relevancy and trustworthiness. When dealing with such problems in non-English ontologies, we can take help from richness of English ontologies. As an example, we can generate data quality tests from English schema data (using approach in Kontokostas et al. \cite{kontokostas}) and use cross-lingual mappings to test quality of non-English instance data. In general, ontology mapping in cross-lingual domain still continues to be a challenge as maintained by  Shvaiko et al. \cite{shvaiko}. The results of OAEI 2013 \cite{oaei13} show that the precision values of the mappings in cross-lingual domain are, in general, poorer as compared to the precision values of mappings in monolingual domain. A similar observation has been made in Fu et al.'s study \cite{bo}.

Most current approaches for cross-lingual ontology mapping (CLOM) are based on machine translation (MT) as pointed out in Fu et al. \cite{bo}. Such approaches first use MT on the source ontology to convert a cross-lingual mapping problem into a monolingual ontology mapping (MOM) problem. Then MOM techniques are used to find mappings to the target ontology. However, a drawback to such an approach (as Fu et al. \cite{bo_mt} points out) is that the mapping performance is critically dependent on the translation step. In recent years, several studies \cite{bo,bo_pseudo,antonis} have attempted to overcome this challenge by devising ways to choose an appropriate machine translation for the labels in the ontologies.

In this paper, our goal is to obtain a $1:1$ cross-lingual \emph{owl:equivalentProperty} correspondence from a given predicate in source ontology to a predicate in the target ontology. We focus only on the limited problem of predicate mapping rather than the broader ontology mapping problem. Yet, many applications can make use of such mappings. For example, cross-lingually mapped predicates can exchange OWL property restrictions from each other. We emphasize that this work on predicate mapping can also be seen as a sub-part of the broader problem of CLOM. Our work, in particular, was motivated by the current scenario in DBpedia instance-data where inter-language links are nearly absent for the predicates but are present only for URIs of individuals. 

Our main contribution in this paper is the Indirect Links Method for cross-lingual predicate mapping. Our system stresses on semantic similarity between mapped predicates rather than lexical similarity. We achieve this mapping by using pre-existing inter-language links of resources connected by the given predicate. Also, we made use of Wikipedia inter-language links. The merit and high-quality of such links have been pointed out in Niu et al. \cite{niu}. Bouma \cite{bouma} has also attempted to use such links for cross-lingual mapping tasks. Additionally, our system uses a lightweight MT+MOM approach as a fallback mechanism.

The rest of the paper is structured as follows. Section 2 describes how our methodology for cross-lingual predicate mapping works in detail. Section 3 provides explanations of the experiment performed for evaluating our methodology. Further, section 4 describes the evaluation results and section 5 concludes our work with pointers to possible future research.

\section{Proposed Methodology for Predicate Mapping}

Our methodology uses two methods, Method 1 and Method 2, which are used in combination with Method 1 as the primary driver. It must be noted that our methodology uses only instance-data for the predicate mapping task. We shall denote the set of all URIs in source ontology (published in language $L_S$) by $S$. Also, we shall denote the set of all URIs in target ontology (published in language $L_T$) by $T$. Both the methods of our methodology take an input predicate $p_S\in S$ and map it to an output predicate belonging to target ontology. We shall denote the output predicate from Method 1 as $p_{T1}$ and that from Method 2 as $p_{T2}$.

Our main contribution lies in the Method 1 (Indirect Links Method) which we shall see in detail next. Further, if the confidence for the returned predicate $p_{T1}$ is low, then we switch to Method 2 as a fallback mechanism and report $p_{T2}$ as output mapping. We denote this finally reported output predicate by $p_T$ and we say that $p_S$ and $p_T$ can be mutually connected by an \emph{owl:equivalentProperty} link.

\subsection{Method 1: Indirect Links Method}
Basic intuition behind this method is that the predicates that link the same pair of resources must have essentially similar meanings. Thus we find the subject-object pairs linked by the given predicate in the source ontology. Then we look for the same or similar subject-object pairs in the target ontology. And finally we extract the predicates that link these subject-object pairs in the target ontology. The idea is represented diagrammatically in Figure 1. A formal description of this algorithm consists of 3 steps and is given below. 

\begin{figure}
\centering
\includegraphics[width=110mm]{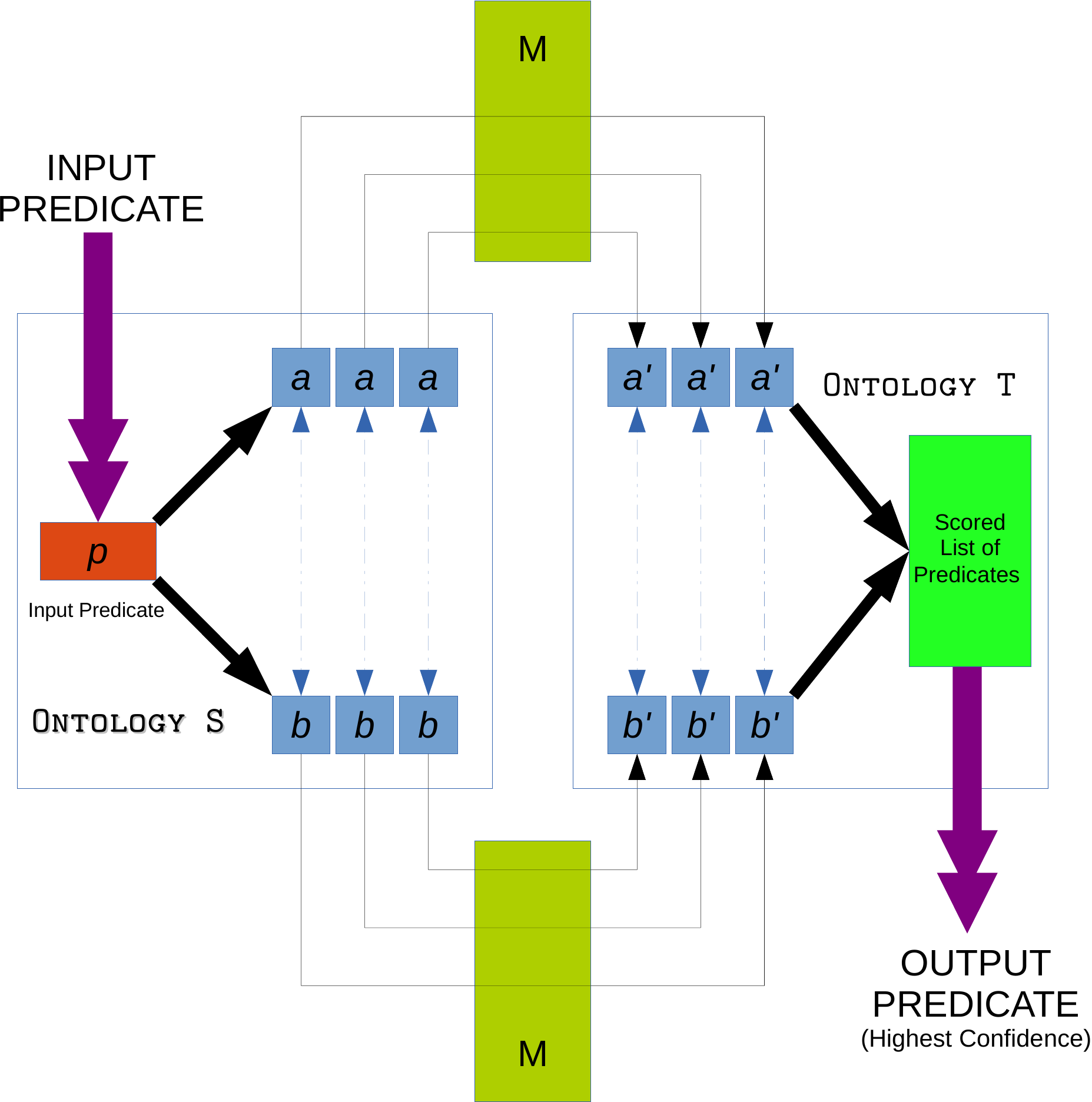}
\caption{Block Diagram for Method 1}
\end{figure}

\subsubsection*{Step 1 : Extracting Subject-Object Pairs}
For the given predicate $p_S\in S$, we obtain set of all the subject-object pairs $(a,b)$ such that triple $(a,p_S,b)$ exists in the source ontology. This is implemented using a SPARQL query as follows:
\begin{lstlisting}
SELECT DISTINCT ?a ?b 
WHERE 
{
	?a <p> ?b .
}
\end{lstlisting}
We will denote this set of subject-object pairs thus obtained by $S_{pairSet}$.

\subsubsection*{Step 2: Mapping Subject-Object Pairs from Source to Target Ontology}
In this step, we obtain cross-ontology mappings for the subjects and objects in the $S_{pairSet}$ obtained in previous step. This step refers to the block M in Figure 1. We treat mapping of each subject or object (which we denote by a dummy variable $u\in S$) as an individual mapping problem where we attempt to find a mapping $u\rightarrow u'$ such that $u'\in T$. Such mappings may be obtained in following possible ways using some pre-existing inter-language link:

\begin{itemize}
\item Using high quality \emph{owl:sameAs} links from $S$ to $T$. 
\item Using string similarity methods on URI $u\in S$ to get corresponding Wikipedia entry title $w_S$ in language $L_S$. Then following the Wikipedia inter-language link to get corresponding Wikipedia entry with title $w_T$ in language $L_T$. Then again employing string similarity techniques to get URI $u'\in T$ corresponding to $w_T$. We noted that such inter-language links are of high quality as also pointed out in Niu et al. \cite{niu}.
\end{itemize}

\noindent
So we obtain the set of subject-object pairs $(a',b')\in T\times T$ such that each pair is mapped from some $(a,b)\in S_{pairSet}$. We shall denote this set by $T_{pairSet}$.

\subsubsection*{Step 3: Relinking the Subject-Object Pairs in Target Ontology}

Let $(a'_i,b'_i)$ be $i^{th}$ subject-object pair in $T_{pairSet}$ and let $P_i$ be the set of all predicates $p'$ such that the triple $(a'_i, p', b'_i)$ exists in the target ontology. We can find this set $P_i$ for a subject-object pair $(a'_i,b'_i)$ using a simple SPARQL query of the following form.

\begin{lstlisting}
SELECT ?p  
WHERE 
{
	<a> ?p <b> .
}
\end{lstlisting}
We take a union of all such $P_i$'s obtained for each $(a'_i,b'_i)\in T_{pairSet}$. Let this union be called $P_T$. Thus, $P_T$ is the set of all the candidate output predicates. In order to select the best predicate suggestion, we shall keep a count (denoted by $c_k$) of the number of subject-object pairs $(a',b')\in T_{pairSet}$ that each predicate $p_k\in P_T$ links. We score each predicate in $P_T$ as follows:
\begin{equation}
score(p_k) = c_k \times log(n_k)
\end{equation}
where $n_k$ is the number of triples containing predicate $p_k$ in the target ontology. We conjectured that $log(n_k)$ acts like a ``Bayesian prior'' and would help reject the deprecated and noisy predicates from occurring in the returned mappings. Confidence for this mapping was calculated as
\begin{equation}
confidence = \frac{score(p_{T1})}{\Sigma_k score(p_k)}
\end{equation}
We chose the predicate $p_k$ with the highest confidence as the output mapping $p_{T1}$. It must be noted this confidence value is calculated only for deciding whether to use Method 2 or not. We do not assign a confidence score for the final output predicate $p_T$.

\subsubsection*{Advantages}
We believe and argue that this method has the following positive features.
\begin{itemize}
\item This method should preserve semantic similarity rather than lexical similarity between mapped predicates. This is because the predicates that link the same pair of resources must have similar meaning. Also, because Wikipedia inter-language links are manually created, they are expected to have high semantic similarity.
\item This method should allow us to find a correspondence even when the number of pre-existing anchor links are sparse/few. This is because multiple subject-object pairs (for any predicate) increase the likelihood of finding at least a few inter-language links.
\item Mapping a large number of subject/object resources for one predicate should cause an averaging/cancelling out of random errors in individual mappings. 
\end{itemize}  

\subsection{Method 2: Google Translate API (on Whole Label) + Edit Distance}
This is a lightweight mapping approach based on standard MT+MOM approach. For a given predicate $p_S\in S$, we first obtain label translations using Google Translate API. Next we use edit distance to find the closest string match of the label translation with the labels of predicates in the target ontology. Tie between many closest matches was broken using a ``prior'' for each predicate. It was calculated as $log(n_k)$ where $n_k$ is the total number of triples containing predicate $p_k$ in the target ontology. 

We use this method as a fallback mechanism when the confidence score for the mapping from the Method 1 is low. We prioritized Method 1 over Method 2 because of greater accuracy of the former when the confidence score was high. 

\section{Evaluation of Proposed Methodology}
To demonstrate the merit and usability of our methodology, we have applied this approach on a pair of following data sets: 
\begin{itemize}
\item {Source Ontology: }Korean DBpedia (\emph{http://ko.dbpedia.org/})
\item {Target Ontology: }English DBpedia (\emph{http://dbpedia.org/})
\end{itemize}

Evaluation was performed by mapping 1000 (out of nearly 16000) predicates which occurred most frequently in the triples of Korean DBpedia data set. This choice of our test-set is justified because 97.5\% of all the triples in Korean DBpedia data set use only these 1000 predicates. We did not extend our test set further because:
\begin{itemize}
\item Remaining predicates form only 2.5\% of the Korean DBpedia data-set and hence are of low significance.
\item Computational effort of testing and human evaluation of the results on remaining data set was high.
\item Low number of triples for such predicates provided lesser anchor links. This underestimated the performance and merit of the algorithm.
\end{itemize}

Evaluation of the predicate mappings obtained from our experiment was performed by two non-author bilingual evaluators with a reasonably high inter-evaluator agreement. Out of 1000 predicate mappings, 200 mappings were rated by both evaluators and a Cohen's Kappa Coefficient \cite{cohen} was calculated, which had a value of 98.38\%. Further, the remaining set of 800 mappings was divided into two equal parts and were given to one evaluator each for evaluation. Evaluators were asked to rate each mapping as one of the following cases:
\begin{enumerate}
\item $p_S \equiv p_T$ i.e. source predicate and target predicate are equivalent.
\item $p_S \models p_T$ i.e. source predicate is a hyponym (i.e., sub-property) of target predicate.
\item $p_T \models p_S$ i.e. source predicate is a hyponym (i.e., sub-property) of target predicate.
\item $p_S \perp p_T$ i.e. source predicate and target predicate are unrelated.
\end{enumerate}  

In our implementation, we invoked Method 2 when the confidence on suggested predicate from Method 1 was lower than a particular threshold $t$. We ran our algorithm by taking this threshold $t$ as 0.1, 0.2 and 0.3. We also ran our algorithm purely using Method 1 only and Method 2 only, respectively. The last case (Method 2 only) served as the baseline for comparison.

In the list below, we mention a few implementation-related intricacies for Method 1. 
\begin{enumerate}
\item In Step 1, the extraction of subject-object pairs in Korean DBpedia was done by executing the corresponding SPARQL query as depicted in Section 2. A SPARQL endpoint for Korean DBpedia version 3.9 was used for the purpose.\\
\item In Step 2, we encountered multiple types of resources while mapping Korean DBpedia subject-object pairs to English DBpedia subject-object pairs. These multiple cases have been given below along with details on how they were tackled.\\
\begin{itemize}

\item \textbf{The subject/object belongs to a standard data-type such as integer, double, date, time etc.}\\ We noted that no mapping from Korean to English was needed because data-typed resources are universal. \\

\item \textbf{The subject/object is a string label}\\
In such cases, we looked for the one-word matches with title labels in Korean Wikipedia and then followed the corresponding Korean-English Wikipedia inter-language link to get an English Wikipedia entry. Then we searched the title label of this Wikipedia entry for one-word matches in the English DBpedia. Thus, we attempt to overcome the language barrier using the Wikipedia inter-language links. Evaluation using a better string matching technique instead of one-word match can be a task for a future work.\\

\item \textbf{Subject/object has a well-formed URI}\\
In such cases, we attempted to look for the inter-language \emph{owl:sameAs} links. It must be noted here that in localised versions of DBpedia (after version 3.7), inter-language \emph{owl:sameAs} links are derived from Wikipedia inter-language links.\\

\end{itemize}
\item In Step 3, retrieval of predicates linking the mapped subject-object pairs was done using the corresponding SPARQL query as depicted in Section 2. A SPARQL endpoint for English DBpedia version 3.9 was used for the purpose.
\end{enumerate}
\noindent

\section{Results}
In this section, we look at the evaluation results for our predicate mapping methodology. In order to put our result in perspective, we shall also compare our results with a baseline matcher based on simple MT+MOM approach. This baseline matcher is based on using Google Translate API on the URI labels in source ontology and then using simple string-based edit distance matcher on this translated ontology.

We argue that this is a fair choice because many recent cross-lingual ontology matchers (as in \cite{bo,bo_pseudo,antonis}) use MT+MOM approaches. We observed that in these matchers, the common choice of machine translation (MT) module is the Google Translate API. Further, for the monolingual ontology mapping (MOM) module, we chose a string-based edit distance matcher because it is simple, lightweight and is still being used by OAEI campaigns for the evaluation of monolingual ontology matchers as in \cite{oaei13}.

One of the demerits of our methodology is that it returns only $1:1$ predicate mappings from source to target ontology. Furthermore, in our current evaluation setup, we have performed only one-way predicate mapping instead of two-way mapping. These issues may be tackled in a future study. Thus, currently, we report only the precision scores and not recall.

Table 1 shows a few examples of mappings given by our system that were rated under different categories.

\begin{CJK}{UTF8}{}
\CJKfamily{mj}

\begin{table} 
\centering 
\begin{tabular}{l c c }
\toprule
Rating & \hspace{30px}Korean DBpedia\hspace{30px} & \hspace{30px}English DBpedia\hspace{30px} \\
& Predicate & Predicate\\\toprule
$p_S \equiv p_T$ & \emph{http://ko.dbpedia.org/property/장르} & \emph{http://dbpedia.org/ontology/genre}\\
\cmidrule(l){2-3}
& \multicolumn{2}{l}{장르 means `Genre' in Korean}\\\midrule
$p_S \models p_T$ & \emph{http://ko.dbpedia.org/property/음반명} & \emph{http://dbpedia.org/property/title}\\
\cmidrule(l){2-3}
& \multicolumn{2}{l}{음반명 means `Album Title'  which is a sub-property of `Title'}\\\midrule
$p_T \models p_S$ & \emph{http://ko.dbpedia.org/property/연도} & \emph{http://dbpedia.org/property/games}\\
\cmidrule(l){2-3}
& \multicolumn{2}{l}{연도 means `Opening Year' which generalises the property `Year of Opening of Games'}\\\midrule
$p_S \perp p_T$ & \emph{http://ko.dbpedia.org/property/후역} & \emph{http://dbpedia.org/property/name}\\
\cmidrule(l){2-3}
& \multicolumn{2}{l}{후역 means `Next Station' and not `Name'}\\\midrule
\bottomrule
\end{tabular}
\vspace{8px}
\caption{Examples of Different Types of Mappings}
\end{table}

\end{CJK}
Table 2 shows the evaluation results for the proposed methodology. Upper half of the table gives the number of mappings under different cases. Lower half gives the precision values of the obtained mappings. Also, Table 3 reports the evaluation results of the baseline system (Method 2 only). 

\begin{table} 
\centering 
\begin{tabular}{l  c c c c} 
\toprule
& \multicolumn{4}{c}{Predicate Mapping Algorithm Used} \\ 
\cmidrule(l){2-5} 	
Rating & \hspace{25px}Method 1\hspace{25px} & \multicolumn{3}{c}{Method 1 + Method 2 (when $confidence\le t$)} \\
\cmidrule(l){2-5} 
& & \hspace{20px}$t=0.10$\hspace{20px} & \hspace{20px}$t=0.20$\hspace{20px} & $t=0.30$\\
\midrule 
$p_S \equiv p_T$ & 500 & 532 & 549 & 557 \\ 
$p_S \models p_T$ & 119 & 119 & 113 & 91 \\ 
$p_T \models p_S$ & 20 & 21 & 18 & 15 \\ 
$p_S \perp p_T$ & 267 & 327 & 317 & 330 \\ 
N/A & 94 & 1 & 3 & 7 \\ 
\midrule 
\midrule 
Precision 1 & \textbf{0.500} & 0.532 & 0.549 & \textbf{0.557}\\
Precision 2 & \textbf{0.639} & 0.672 & \textbf{0.680} & 0.663\\
\bottomrule 
\end{tabular}
\vspace{8px}
\caption{Evaluation Results of the Proposed Methodology}
\end{table}

\begin{table} 
\centering 
\begin{tabular}{l  c} 
\toprule
Rating\hspace{25px} & Number of Mappings  \\
\midrule 
$p_S \equiv p_T$ & 436  \\ 
$p_S \models p_T$ & 25  \\ 
$p_T \models p_S$ & 10 \\ 
$p_S \perp p_T$ & 523  \\ 
N/A & 6 \\ 
\midrule 
\midrule 
Precision 1 & \textbf{0.436} \\
Precision 2 & 0.471 \\
\bottomrule 
\end{tabular}
\vspace{8px}
\caption{Evaluation Results of the Baseline Methodology}
\end{table}

We have calculated 2 kinds of precision scores based on whether partially correct mappings of type $p_S \models p_T$ or $p_T \models p_S$ are considered or not. It should be noted that mappings rated as $p_S \models p_T$ or $p_T \models p_S$ cannot be thoughtlessly disregarded because it reflects that input and output predicates had some semantic similarity, prompting us to report precision scores that take these partially correct mappings into account. These two kinds of precision values are described below:
\begin{itemize}
\item Precision 1 refers to fraction of total mappings that were rated $p_S \equiv p_T$.
\item Precision 2 refers to fraction of total mappings that were rated $p_S \equiv p_T$ or $p_S \models p_T$ or $p_T \models p_S$.
\end{itemize}

\noindent
We summarise some of the inferences as follows.
\begin{enumerate}
\item Proposed methodology (having precision 0.56) is an improvement over the baseline system (having precision 0.44) as evidenced by the 22\% higher precision of the former.
\item Proposed methodology is able to identify many possible mappings with partial semantic linkage. This is evidenced by the high number of mappings rated $p_S \models p_T$ and $p_T \models p_S$ furnished by our methodology as compared to the baseline system. This supports our hypothesis that our method stresses on semantic similarity rather than lexical similarity. Thus, taking partially correct mappings into consideration, we see a huge improvement in precision (of nearly 45\%) from 0.47 (for baseline system) to 0.68 (for proposed methodology).
\item A drawback of Method 1 is that nearly one in ten mappings could not be found. This generally occurred due to the lack of Wikipedia entry for the  subjects/objects in Korean or English DBpedia. Also, it occurred due to the lack of Wikipedia inter-language links. Also, it was manually observed that Method 1 returned more mappings rated $p_S \perp p_T$ when confidence was low as compared to when it was high. These drawbacks have been successfully overcome by using Method 2 as a fallback mechanism. Setting confidence threshold $t=0.3$ yields the maximum precision value of 0.56 (disregarding partially correct mappings). This is an improvement over the precision value of 0.50 (disregarding partially correct mappings), which was obtained by using only Method 1.  
\end{enumerate}

\section{Conclusion and Future Work}
We have presented a successful ad-hoc system for cross-lingual predicate mapping. We have successfully demonstrated that our proposed methodology outperforms the baseline system. Further, we have demonstrated that our system stresses on semantic similarity rather than lexical similarity as our system outperforms the baseline system in furnishing partially correct mappings.

This work still leaves a lot of scope for future work. Currently, the evaluation has been performed on a limited data set between just one pair of ontologies. So performing evaluation on more pairs of ontologies can shed more lights on merits and demerits of our methodology. 

One of the key issues with this work is that the algorithm returns only $1:1$ predicate mappings from source to target ontology. Thus, algorithm misses out on other possible predicates in the target ontology that might capture the same role. Hence, this work may be extended to include this additional aspect and to get $1:n$ mappings.

There can be various ways is which the current implementation may be improved. One possibility is to explore other ways to cross the natural language barrier instead of using Wikipedia inter-language links. Another possibility is to have greater and better support for different data-types for cross-lingual mapping of subjects and objects. Additionally, we can have a more sophisticated measure of confidence for the predicate suggestion because current metric is based on a simple ratio-based heuristic. A future work may also attempt on distinctly identifying not just $p_S \equiv p_T$ mappings but also mappings of type $p_S \models p_T$ and $p_T \models p_S$. Notwithstanding these limitations and future research issues, the current study provides a novel approach and important insights towards overcoming the disparity between cross-lingual ontologies.


\begin{thebibliography}{5}
\bibitem{bizer} Bizer, C., T. Heath, and T. Berners-Lee. ``Linked data-the story so far." International journal on semantic web and information systems 5.3 (2009): 1-22.
\bibitem{shvaiko} Shvaiko, P. and J. Euzenat. ``Ontology matching: state of the art and future challenges." Knowledge and Data Engineering, IEEE Transactions on 25.1 (2013): 158-176.
\bibitem{niu} Niu, X., et al. ``Zhishi.me-weaving Chinese linking open data." The Semantic Web–ISWC 2011. Springer Berlin Heidelberg, 2011. 205-220.
\bibitem{kontokostas} Kontokostas, D., et al. ``Test-driven evaluation of linked data quality." Proceedings of the 23rd international conference on World Wide Web. International World Wide Web Conferences Steering Committee, 2014.
\bibitem{bo} Fu, B., R. Brennan, and D. O’Sullivan. ``A configurable translation-based cross-lingual ontology mapping system to adjust mapping outcomes." Web Semantics: Science, Services and Agents on the World Wide Web 15 (2012): 15-36.
\bibitem{bo_mt}Fu, B., R. Brennan, and D. O’Sullivan. ``Cross-lingual ontology mapping–an investigation of the impact of machine translation." The Semantic Web. Springer Berlin Heidelberg, 2009. 1-15.
\bibitem{bo_pseudo} Fu, B., R. Brennan, and D. O’Sullivan. ``Using pseudo feedback to improve cross-lingual ontology mapping." The Semantic Web: Research and Applications. Springer Berlin Heidelberg, 2011. 336-351.
\bibitem{antonis} Koukourikos, A., P. Karampiperis, and G. Stoitsis. ``Cross-Language Ontology Alignment Utilizing Machine Translation Models." Metadata and Semantics Research. Springer International Publishing, 2013. 75-86.
\bibitem{bouma} Bouma, G. ``Cross-lingual Dutch to English alignment using EuroWordNet and Dutch wikipedia." Proceedings of the 4th International Workshop on Ontology Matching, CEUR-WS. Vol. 551. 2009.
\bibitem{oaei13} Grau, B. C., et al. ``Results of the ontology alignment evaluation initiative 2013." Proc. 8th ISWC Workshop on Ontology Matching (OM). 2013.
\bibitem{cohen} Cohen, J. "A Coefficient of Agreement for Nominal Scales." Educational and Psychological Measurement 20(1), 1960, 37-46.
\end{thebibliography}
\end{document}